\documentclass[letterpaper, 10 pt, conference]{ieeeconf}  % Comment this line out if you need a4paper

\IEEEoverridecommandlockouts                              % This command is only needed if 
\usepackage{graphicx}
\graphicspath{{figs/}}
\usepackage{amsmath}
\usepackage{amssymb}
\usepackage{algorithm}
\usepackage{algpseudocode}
\usepackage{times}
\usepackage[position=bottom]{subfig}
\usepackage{siunitx}
\newcommand{\seclab}[1]{\label{section:#1}}
\newcommand{\secref}[1]{Section \ref{section:#1}}
\newcommand{\figlab}[1]{\label{figure:#1}}
\newcommand{\figref}[1]{Fig.\ref{figure:#1}}

\title{\LARGE \bf
3D Object Segmentation for Shelf Bin Picking by Humanoid \\ with Deep Learning and Occupancy Voxel Grid Map
}

\author{Kentaro Wada$^{1}$ and Masaki Murooka$^{1}$ and Kei Okada$^{1}$ and Masayuki Inaba$^{1}$% <-this % stops a space
  \thanks{K. Wada, M. Murooka, K. Okada, and M. Inaba are with the Department of Mechano-Informatics. %
    The University of Tokyo, 7-3-1 Hongo, Bunkyo-ku, Tokyo 113-8656, Japan. %
    wada@jsk.imi.i.u-tokyo.ac.jp.%
  }
}

\begin{document}

\maketitle
\thispagestyle{empty}
\pagestyle{empty}

\begin{abstract}
Picking objects in a narrow space such as shelf bins is an important task
for humanoid to extract target object from environment.
In those situations, however,
there are many occlusions between the camera and objects,
and this makes it difficult to segment the target object three dimensionally
because of the lack of three dimentional sensor inputs.
We address this problem with accumulating segmentation result with multiple camera angles,
and generating voxel model of the target object.
Our approach consists of two components: first is object probability prediction
for input image with convolutional networks,
and second is generating voxel grid map which is designed for object segmentation.
We evaluated the method with the picking task experiment for target objects in narrow shelf bins.
Our method generates dense 3D object segments even with occlusions,
and the real robot successfully picked target objects from the narrow space.

\end{abstract}

% 1. Introduction
\section{INTRODUCTION}

\begin{figure}[tbhp]
  \centering
  \hspace{-4mm}
  \begin{tabular}{c}
    \subfloat[]{
      \includegraphics[width=0.27\textwidth]{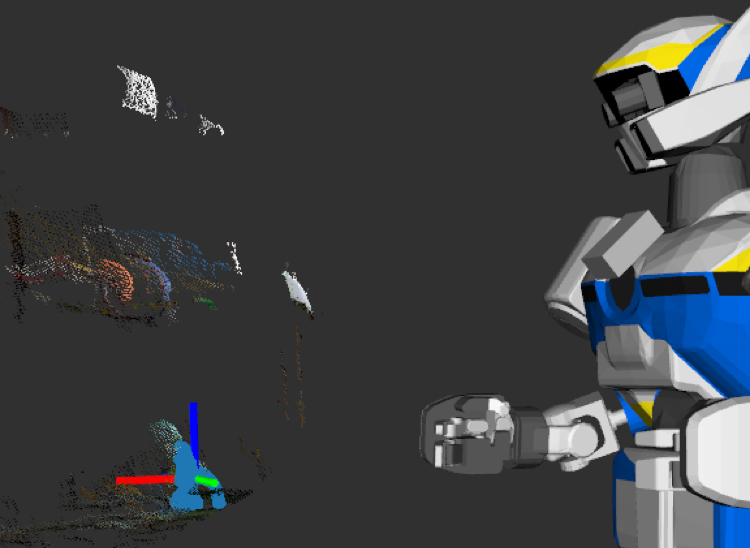}
      \figlab{whatisthis1}} \\
    \subfloat[]{
      \includegraphics[width=0.27\textwidth]{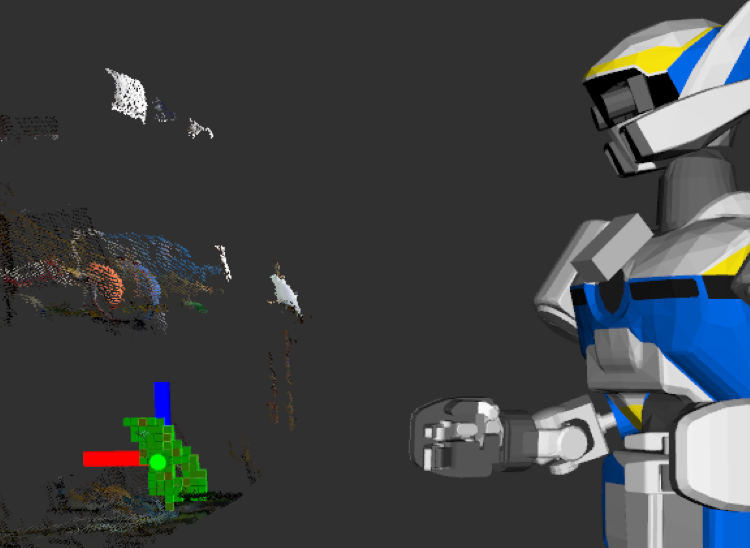}
      \figlab{whatisthis2}}
  \end{tabular}
  \hspace{-7mm}
  \begin{tabular}{c}
    \subfloat[]{
      \includegraphics[width=0.18\textwidth]{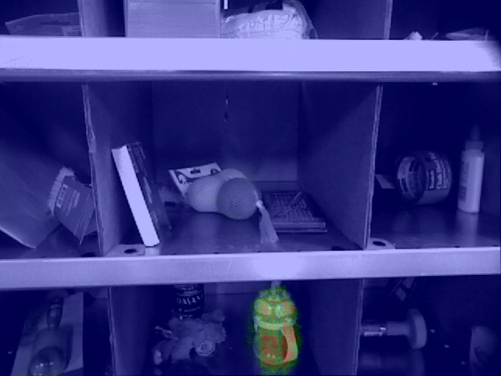}
      \figlab{whatisthis3}} \\
    \subfloat[]{
      \includegraphics[width=0.18\textwidth]{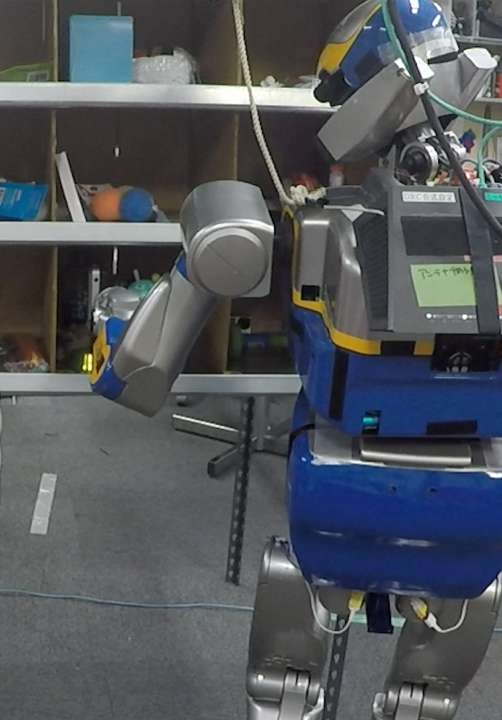}
      \figlab{whatisthis4}}
  \end{tabular}
  \caption{Shelf bin picking by humanoid robot.
    \newline
    \footnotesize{
      Each image is a result of segmentation with usual method
      (\figref{whatisthis1}),
      that with our method (\figref{whatisthis2})
      robot camera view at recognizing (\figref{whatisthis3})
      with red region representing high probability of the target,
      and picking behavior by real robot (\figref{whatisthis4}).
      The target object, green cup, is placed at deep position in the bin,
      and it is hard to segment it three dimentionally
      with usual method because of the occlusions.
      Our method segments the target object densely compared to the usual method
      refering the point cloud.
    }
  }
  \label{figure:what_is_this}
\end{figure}

% ロボットピッキングは取り組まれて解かれている部分はあるが，
% 閉鎖空間などのオクルージョンのある場合では解かれていない．
Robotic picking capability, which is fundamental for manipulation of objects,
has progressed in recent years so that which can be applied for various objects.
But it is still difficult to pick objects
in a narrow space such as shelf bin picking environment as shown in \figref{what_is_this}
with uncertainty of object pose, brightness, and occlusions.
Even with the recent progress of machine 2D object segmentation technology with deep learning
\cite{pascal-voc-2012} \cite{long_shelhamer_fcn},
the lack of three dimentional sensor inputs because of occlusions is still the problem
on segmenting 3D object for robotic manipulation.
% In particular for picking task to manipulate target object,
% robustness to these uncertainty is needed to segment strictly the target object.

% 閉鎖空間ではオクルージョンがあることがタスクを困難にしている．
Compared to the picking task in an environment
where previous studies were conducted like a tabletop,
picking task in a narrow space has difficulty of occlusions.
This is because the objects is located inside the narrow space like bins
shown in \figref{what_is_this}, and the appropriate camera angles
for object localization is restricted
by the wall and roof parts as well as by the objects themselves.
In general, it is hard task for robot to pick target object with many occlusions,
because the lack of 3D information of the object
makes both localization and size estimation difficult,
which are crucial for generating picking motion.
% 閉鎖空間(closed space)におけるピッキングタスクにおいては，テーブルトップなどの通常の環境に比べて
% オクルージョンが多いという特徴がある．
% これは，物体が\figref{what_is_this}に示す棚のような奥まった領域に置かれており，
% 通常の環境でも存在するような物体同士のオクルージョンだけでなく，
% 壁や屋根部分によって物体を可視可能な視点が限定されることによる．
% 一般にオクルージョンが多い環境では，隠れのある部分で点群情報が欠けてしまい，
% ヒューマノイドのピッキングタスク実現において不可欠となる目標物体の三次元位置同定は困難である．

% % 閉鎖空間では物体とのインタラクションなどを行うのもリスキーである．
% さらに，物体に対して作用することで物体姿勢を変更し，
% 認識の容易な状態へと変化させるような手法も考えられるが，
% Humanoidsによるこのようなインタラクションは閉鎖空間においてはRiskyである．

% 見回しによる三次元モデル生成と物体セグメンテーションが必要である．
In this study,
we propose a method for three dimentional object segmentation in a condition
with difficulty on acquiring three dimensiontal sensor inputs, point cloud.
Our proposing method consists of 2D object segmentation and 3D voxel grid mapping.
In addition to the standard method to convert 2D segmentation to 3D voxels,
which uses the correspondence between image and point cloud
(upper left in \figref{what_is_this}),
we accumulate each segmentation result in different viewpoints as a 3D map
(lower left in \figref{what_is_this}).
Our method enables the robot to acquire
a dense three dimentional information on segmenting the target object,
and expands the scope of activity of humanoids for manipulation in narrow space.
% 本研究では，このように物体の三次元情報である点群をセンシングしにくいような状況において
% 目標物体の三次元位置同定を行う手法を提案する．
% 提案手法は二次元画像上での物体セグメンテーションと三次元マッピングからなる．
% 二次元画像上での物体セグメンテーションを画像と点群の対応関係から三次元化するという
% 一般的な手法に加えて，物体三次元的存在を三次元マップとして表すことでオクルージョンのある環境
% においても物体の3次元形状を密に獲得することが可能となる．
% 本手法を基に，ヒューマノイドの閉鎖空間におけるピッキングタスクを
% 可能にすることが本研究の目標である．

% 2. Semantic Object Localization for Humanoids Picking
\section{3D Object Segmentation \\ for Picking by Humanoid}

% ## 関連研究
\subsection{Related Works}

% ### 物体ピッキング
Robotic picking is one of the fundamental robotic manipulation behavior,
and studied in a many previous works.
% ロボットによる物体ピッキングタスクに関する研究としては多くの試みが成されている．
% それらは複数の物体が写った画像入力から把持点を決定する．
% しかしこれらは物体の識別を考慮せずに物体を把持する機構であり，
% 物体をセグメンテーションした結果を入力と出来ない場合など，
% 目標物体の位置同定機構との親和性が低い場合がある．
In previous works, however,
the robot workspace is mostly collision and occlusion free,
and they conduct the picking task in a situation with
given location of target object \cite{Tegin:SimGrasp},
no target in various objects \cite{DBLP:journals/corr/LevinePKQ16}
\cite{DBLP:journals/corr/PintoG15},
and in situation with single object on a tabletop \cite{Kei:HRP2}.
Even in a previous work for picking task in a narrow space,
the target object is located at the front side of the shelf bin
\cite{RBO},
whose 3D information can be more easily sensed by RGB-D sensor:
\figref{depth_in_narrow_space_002} is the harder situation than \figref{depth_in_narrow_space_001}
with objects located in inside back of the shelf bin.
In \figref{depth_in_narrow_space_002}, the objects are occluded by the part of the shelf,
and the depth of transparent part of the cup is not sensed because of the camera angle.
% さらにロボットの作業空間として棚の中などの壁が迫っているような閉鎖空間での作業を行った例は少なく，
% \cite{RBO}
% 既存の研究でもカメラによってセンシングしやすいような棚の入り口付近に物体を設置してあるような
% 状況でのタスク実現が主である．
% \cite{Kei}
In this study, we tackle the difficulty of picking task execution
in a condition with occlusions of target object and lack of 3D sensor inputs.
% 本研究では，目標物体が複数物体の隣接やオクルージョン下にある状況における
% ピッキング作業実現の困難性に取り組み，ヒューマノイドのピッキング作業可能範囲を拡大する．

\begin{figure}[htbp]
  \centering
  \subfloat[]{
    \includegraphics[width=0.45\textwidth]{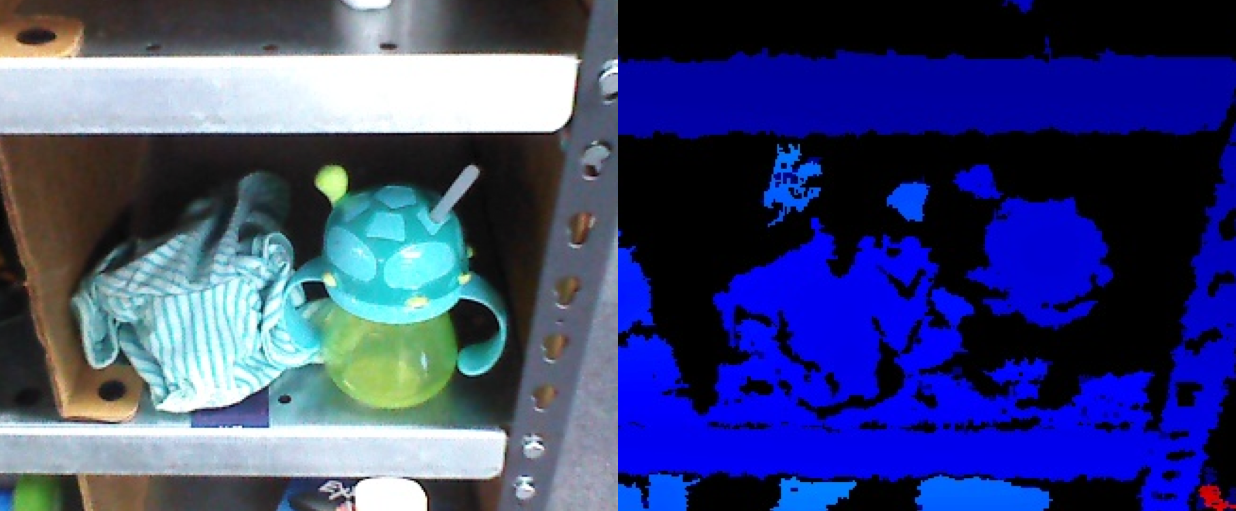}
    \figlab{depth_in_narrow_space_001}
  }

  \subfloat[]{
    \includegraphics[width=0.45\textwidth]{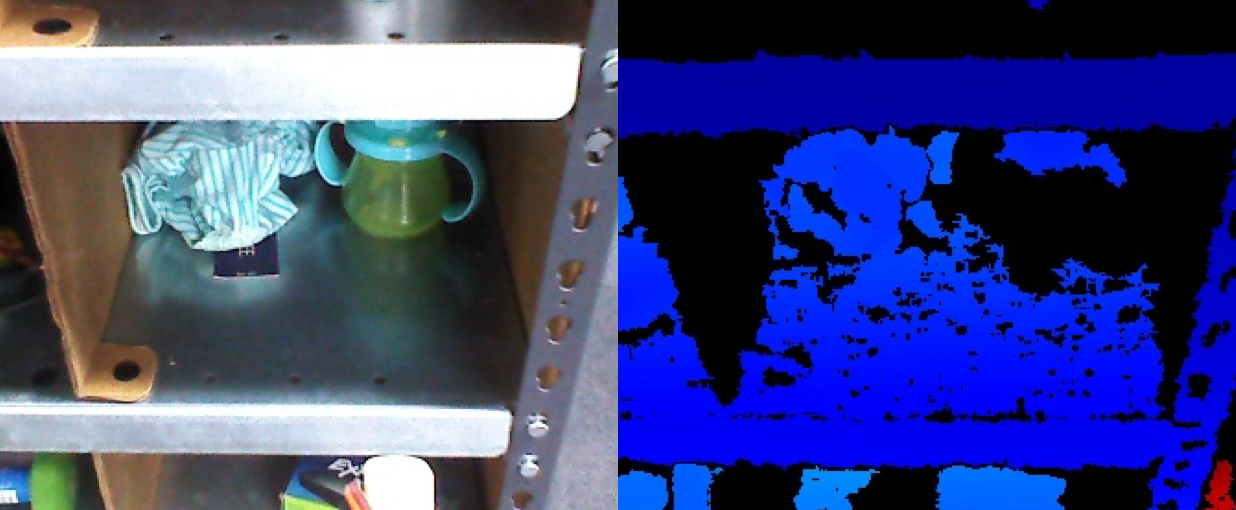}
    \figlab{depth_in_narrow_space_002}
  }
  \caption{3D information of objects in a narrow space.
    \newline
    \footnotesize{
      It shows RGB (left) and depth image (right).
      \figref{depth_in_narrow_space_001} shows
      the situation where objects are located in front,
      and \figref{depth_in_narrow_space_002} shows one with objects in back.
    }
  }
  \figlab{depth_in_narrow_space}
\end{figure}

% ### 物体セグメンテーションに関する研究: 2D-3D, Mapping-Maplessの対比
As the previous works on object segmentation,
it is tackled using a large-scale dataset
with thousands of images and dozes of object classes
\cite{pascal-voc-2012}.
% 物体セグメンテーションに関する研究としては，
% 二次元画像を用いたものとして過去にPascal VOCデータセットなどの
% 数千枚画像で数十クラスの問題設定において取り組まれている．
% \cite{pascal}
% これに対して，ロボットが実環境においてタスクを行うためには3次元的な物体位置同定が必要となり，
% 一般には二次元画像と三次元点群の対応関係から三次元的なセグメンテーション結果とする方法が取られる．
% \cite{RBO}
In contrast to this 2D segmentation work,
segmenting three dimensionally is required for robot to localize object
in order to conduct tasks in real world.
The usual approach for this is registering the 2D segmentation result to the 3D point cloud
with correspondence relationship between image and point cloud \cite{RBO}.
In another work \cite{Stckler2012SemanticMU}, mapping-based segmentation is tackled
using probabilistic 2D image segmentation and updating voxels with a Bayesian framework.
That work was tackled in order to improve the segmentation accuracy on 2D image by
accumulating the probability in multiple views.
Our approach also includes this component to accumulate 2D segmentation result
to acquire dense 3D information of the target object in an environment with occlusions.
And the main difference in the approach between \cite{Stckler2012SemanticMU} and our work is
the 2D segmentation method: random decision forests in previous work
and deep convolutional network in our work.
\subsection{Proposed Method}

\begin{figure}[htbp]
  \centering
  \includegraphics[width=0.48\textwidth]{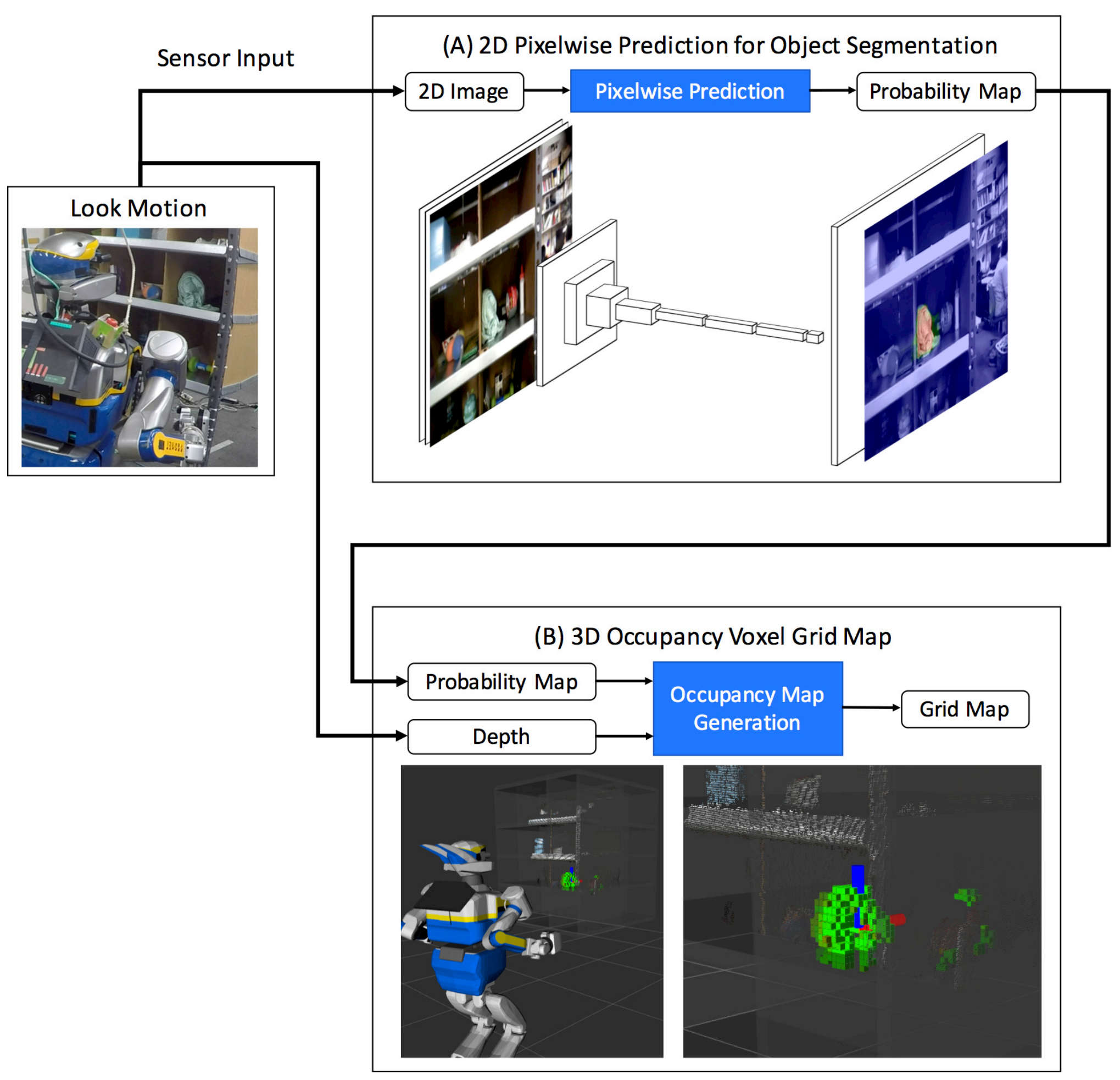}
  \caption{Proposed system for 3D object segmentation. \newline
    \footnotesize{
      The main component of this system is the pixelwise object probability prediction
      based on convolutional network, and the voxel grid map generation
      with updating the occupancy probability in each grid.
    }
  }
  \figlab{proposed_system}
\end{figure}

% ### 提案手法
Our proposing method for object segmentation consists of 2 components:
first is the object probability prediction $I^{proba}_t = f(I_t)$,
which predicts pixelwise label occupancy probability for candidate objects $I^{proba}_t$
at time $t$ from input image $I_t$,
and the second is 3D voxel grid map generation $M_t = g(M_{t-1}, I^{proba}_t, C_t)$
which updates the voxel grid map $M_{t-1}$
with the probability map $I^{proba}_t$ and point cloud $C_t$.
In the second component, the probability 2D prediction result is registered to the point cloud
to generate the 3D probability prediction result,
and the 3D map is used to update the voxel grid map.
% 本研究で提案するセグメンテーション手法は以下の2つの要素からなる．
% 一つは時刻$t$における二次元画像$I^_t$から各ピクセル毎に候補物体群の
% 各クラスの占有確率$I^{probaAll}_t$を出力するもので，
% 畳み込みニューラルネットワーク$F$
% によって画像入力から同サイズの確率画像を推定する．
% もう一つは目標物体の確率画像$I^{probaTarget}_t$と三次元点群$C_t$
% を入力として三次元的な確率マップ$M_t$として蓄積するもので，
% 二次元の確率マップを三次元点群にレジストレーションすることによって三次元点群の点ごとに
% 確率を獲得する．
% 獲得した三次元的な確率マップをグリッド状に分割されたマップにおいて
% 各グリッドを目標物体が占有しているか否かの確率という確率として更新する．

We construct system shown in \figref{proposed_system}
by integrating our 3D segmentation method with looking around motion of humanoid.
By accumulating the object probability prediction result three dimensionally as a map,
we realize the target object segmentation in a narrow space.

% ヒューマノイドの見回し動作において本手法を適用することで
% \figref{proposed_system}に示すようなシステムを構築する．
% 異なる視点における確率画像と三次元点群をマップとして蓄積することで，
% 閉鎖空間において目標物体の三次元的なセグメンテーションを実現する．

In the following sections, we introduce the method to predict object probability
from input image in \secref{fcn}, and the map generation method in \secref{octomap}.
In \secref{experiment}, we validate the efficiency of our proposed method
with a shelf bin picking task execution in the real world.

% 後に本論文では，
% 第3章で畳み込みニューラルネットワークによって画像入力から確率画像を出力する手法に関して，
% 第4章で確率画像と点群から三次元確率マップを生成する手法に関して述べる．
% 第5章ではこれらによって目標物体の三次元セグメンテーションを行い閉鎖空間にてピッキングを行う
% 実機実験を行い，本手法を有効性を検証する．

% 3. 2D Object Segmentation with Convolutional Networks
% ## 畳み込みニューラルネットワークによる物体セグメンテーション
\section{2D Object Segmentation \\ with Fully Convolutional Networks}  \seclab{fcn}

In this section, we introduce the function $I^{proba}_t = f(I_t)$ which receives
RGB image $I_t$ as the input and outputs object probability image $I_t^{proba}$.
We construct deep neural network for this function,
and we describe the network architecture and the way of training in the following subsections.
% RGB画像$I_t^{raw}$を入力として物体の確率画像$I_t^{probaAll}$を出力とするような
% 関数$I_t^{probaAll} = F(I_t^{raw})$を設計する．
% 本章では，Deep Learningを用いてこのような関数$F$を得る手法に関して，
% ネットワーク構造と学習方法の二点述べる．

% ### ネットワーク構造
\subsection{Network Architecture}

\begin{figure}[htbp]
  \centering
  \includegraphics[width=0.48\textwidth]{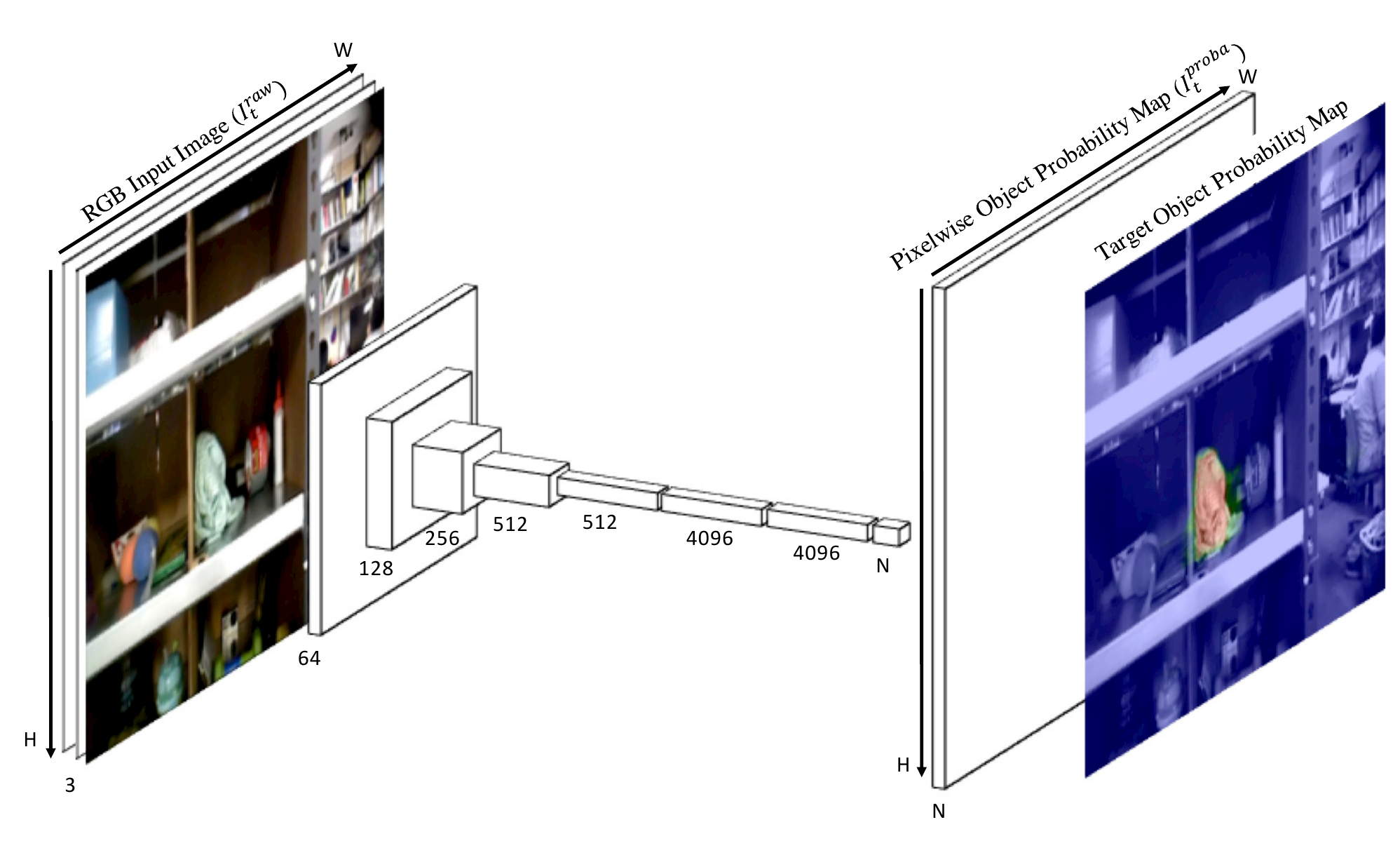}
  \caption{Fully Convolutional Networks architecture. \newline
    \footnotesize{
      Our network architecture for the 2D object segmentation with 40 object labels.
    }
  }
  \label{figure:network_architecture}
\end{figure}

% FCNは物体セグメンテーションにおいて有効なモデル
For object segmentation with deep learning,
Fully Convolutional Networks (FCN) architecture is usually used \cite{long_shelhamer_fcn}.
In this network model, the convolutional operation is applied to the input image with half resizing
in max pooling layers, and the inverse operation of convolution (deconvolution) is applied and the
hidden layers output is resized to the same size with the input image as the output layer.
The network architecture is shown in \figref{network_architecture}: the number in the figure
represents the number of channels, the box size represents the image size,
$H$ represents input image height,
$W$ represents width, $N$ represents the number of object labels.
In general, object segmentation is a problem to assign object labels to each pixels, and
it is requested to output label image, whose size is same as the input image and
each pixel has object label values.
In previous works \cite{pascal-voc-2012} \cite{long_shelhamer_fcn},
convolution neural networks extract features from image with keeping the 2D construction,
and it is confirmed that FCN is the powerful architecture for this kind of task,
in which both input and output are image.
% 一般に二次元画像を入力とした物体セグメンテーションとしては，
% 畳み込みニューラルネットワークを多段に積み上げたFully Convolutional Networks (FCN)
% というモデルが用いられる．
% このモデルは\figref{network_architecture}に示すように
% 入力画像を畳み込み処理を適用しながら小さくし，
% 出力層で畳み込み処理の逆処理を行い(Deconvolution)画像サイズを入力画像と同じサイズに戻す処理を行う．
% 図において数字は各層の出力のチャネル数，直方体の大きさは画像サイズを表し，
% $H$は入力画像の高さ，$W$は入力画像の幅，$N$は物体ラベル数を表している．
% 物体セグメンテーションは各ピクセルにどの物体の値(Label)を割り振るかという問題で，
% センサからの画像を入力として各ピクセルにLabelが入ったラベル画像を出力することが求められる．
% 畳み込みニューラルネットワークは画像の二次元構造をを保ったまま特徴量抽出が可能であり，
% 先行研究からFCNは入出力が共に画像となっているような問題に関して有効であることがわかっている．

% 出力を確率マップとした
In this study,
we also used the similar FCN architecture shown in \figref{network_architecture}
for object segmentation as the previous work \cite{long_shelhamer_fcn}.
The network consists of 16 convolution layers, 5 max pooling layers, and 1 deconvolutional layer,
and as the activation function Relu \cite{icml2010_NairH10} is used.
We used the softmax function to output probability map $I^{proba}_t$
from the output of deconvolutional layer $I^{deconv}$
for $i = 0 \cdots H$ and $j = 0 \cdots W$ as below:

% 本研究でも先行研究と同様のネットワーク構造を用い，16層の畳み込み層と5層のPooling層からなる．
% 活性化関数としては主にRelu関数を用いたが，最終層の出力を確率マップとするために，
% Deconvolution層の出力$I^{deconv}$に対してSoftmax関数%\cite{softmax}
% を適用した．

\begin{eqnarray}
  I^{proba}_{ij} = \sigma (I^{deconv}_{ij}) \\
  \sigma(I^{deconv}_{ij}) = \frac{exp(I^{deconv}_{ij})}{\sum^N_{k=0}exp(I^{deconv}_{ijk})}
\end{eqnarray}

% ### 学習方法
\subsection{Training the Network with Dataset}

\begin{figure}[htbp]
  \centering
  \includegraphics[width=0.48\textwidth]{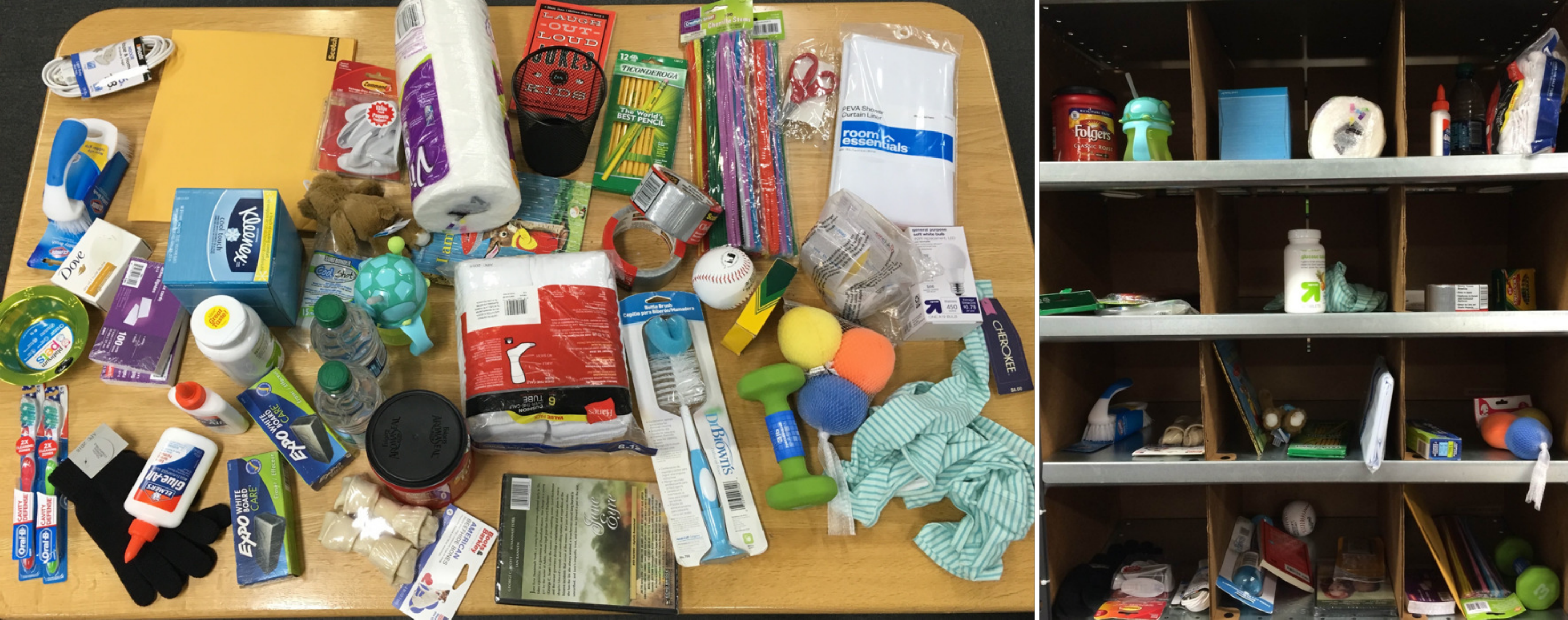}
  \caption{Objects and environment for the picking sernario. \newline
    \footnotesize{
      We use 39 items for the experiment, and a shelf which has 12 bins.
      The number of object classes is larger than that of previous works,
      but the environment where objects exist is limited in the shelf.
    }
  }
  \label{figure:segmentation_problem}
\end{figure}

% 40クラスのセグメンテーション問題であり，そこそこの難しさ．
We handle items shown in \figref{segmentation_problem}, so the
whole object labels for segmentation is 40 labels: 39 labels for the items and 1 background label.
Compared to the previous researches \cite{pascal-voc-2012} \cite{long_shelhamer_fcn},
which handle 21 object classes, the number of that we are handling is larger.
But in our picking senario the objects are only located inside the shelf, and the location
where objects is located is limited.
% 本研究では\figref{segmentation_problem}に示す39個の物品ラベルと
% 1個の背景ラベルの計40クラスの物体セグメンテーション問題を扱う．
% 先行研究で扱っていたような21クラスの物体セグメンテーションと比べると物体クラス数では多い一方，
% 物体の置かれている状況としては棚の中に置かれている場合のみで限定されているという特徴がある．

\begin{figure}[htbp]
  \centering
  \includegraphics[width=0.48\textwidth]{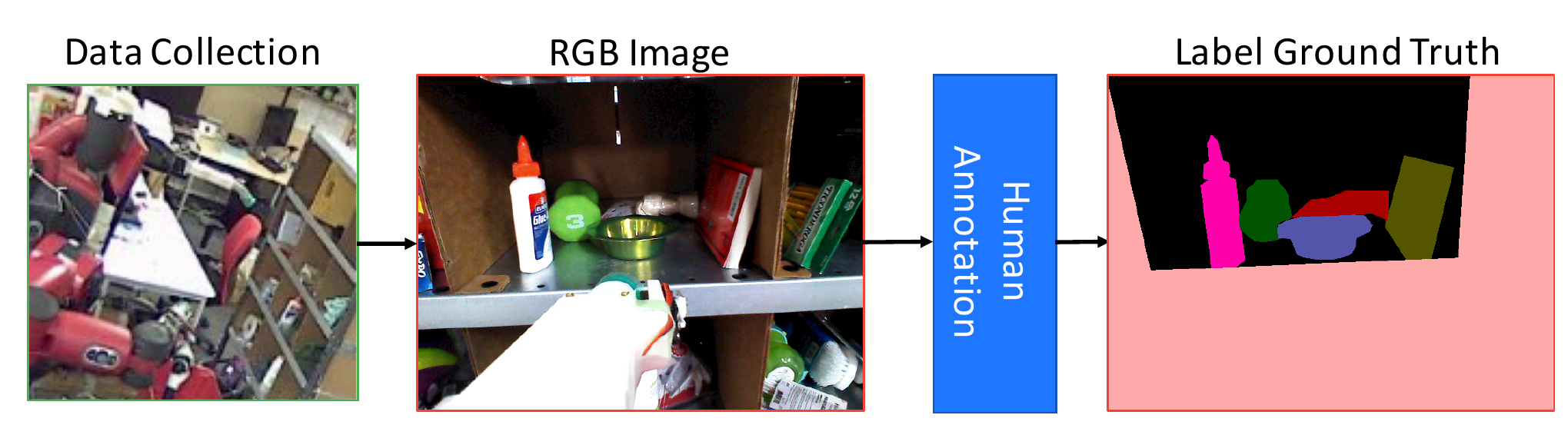}
  \caption{Process for generating image segmentation dataset. \newline
    \footnotesize{
      The process consists of automated image collection by a robot,
      and human annotation.
    }
  }
  \label{figure:dataset}
\end{figure}

% 棚の中での物体セグメンテーション問題としてどのようなデータセットを作成したか．
Considering the restriction of the objects location,
we collected dataset with images in which the objects are located inside the shelf bins
as shown in \figref{dataset}.
The dataset generation process consists of the image collection and human-handed annotation,
and we automated the image collection by using a robot which had RGB-D camera sensor on its hand.
The size of the dataset is 218 sets of sensor image
and label image generated with annotation by human.
We splitted the dataset in 8:2 for training and validation, and used them
for network training and evaluation of segmentation accuracy.
% 本研究では，ピッキングタスクにおける物体セグメンテーションに限定し，
% \figref{dataset}に示すような棚の中に物体が置かれた状況のデータセットを作成した．
% データセットの作成はロボットによる生画像の収集と人手によるアノテーションからなり，
% 画像の収集は手先にRGB-Dカメラのついたロボットを利用することで自動化した．
% データセットは生画像とラベル画像を一組として218組作成した．
% これらを8:2に学習用とテスト用とに分割し，それぞれをネットワークの学習と汎化性能の評価に用いた．

\begin{figure}[htbp]
  \centering
  \hspace*{-6mm}
  \includegraphics[width=0.55\textwidth,trim={5mm 0 5mm 5mm},clip]{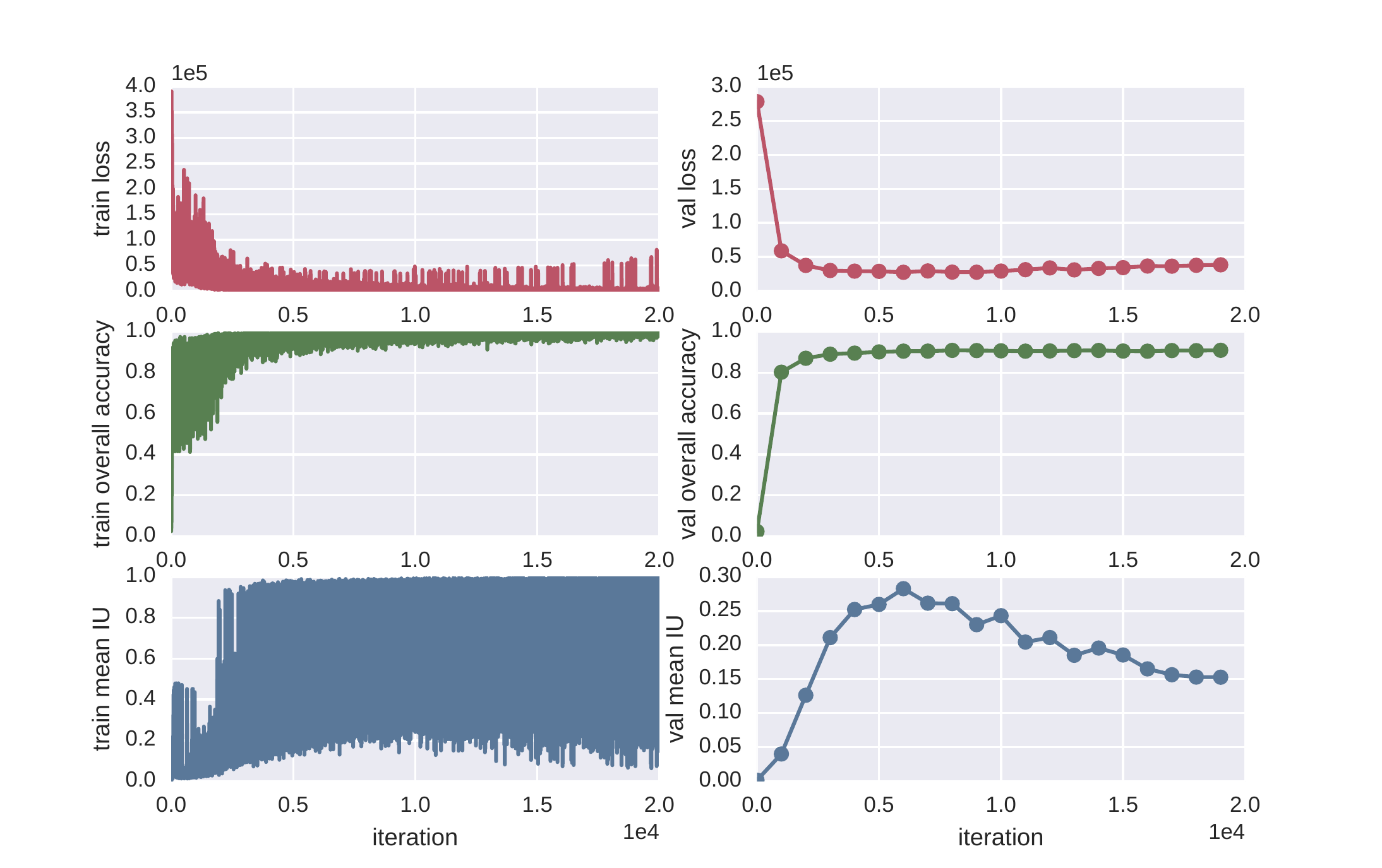}
  \caption{Learning curve on training of the segmentation network. \newline
    \footnotesize{
      Learning curve for loss, pixelwise accuracy,
      and mean IU for both training (left) and validation (right) dataset.
      The x axis in all figures represent training iterations.
    }
  }
  \label{figure:learning_curve}
\end{figure}

% どのように学習して，結果はどうだったか．
Following the previous work
we used the parameters of VGG16 network \cite{DBLP:journals/corr/SimonyanZ14a}
to initialize parameters of convolutional layers in FCN,
by fine-tuning of the object recognition network for object segmentation
as its efficiency is reported in previous work \cite{NIPS2014_5347}.
% 畳み込み層の重みの初期値としてはImagenetのデータセットにおいて学習済みのVGG16ネットワークを用い，
% それらをファインチューニングする形で学習を行った．
The loss function is the softmax cross entropy of deconvolutional layer's output
and ground truth label image,
and the optimization is conducted by Adam \cite{DBLP:journals/corr/KingmaB14}
with parameters of step size: $\alpha=\num{1e-5}$,
and exponential decay rates for the moment estimates: $\beta_1=0.9$, $\beta_2=0.99$.
% 学習におけるロス計算はDeconvolution層の出力$I^{deconv}$と正解ラベル画像$I^{gt}$のSoftmax Cross Entropy誤差とし
% 最適化はAdamに対するパラメータ$\alpha$値$1e-5$と$\beta$値$0.9$によって行った．
We compute the 2D segmentation accuracies defined below:
\begin{itemize}
  \item pixelwise accuracy: $\sum_i n_{ii} / \sum_i t_i$
  \item mean IU: $(1 / n_{cl}) \sum_i n_{ii} / (t_i + \sum_j n_{ji} - n_{ii})$
\end{itemize}
where $n_{ij}$ is the number of pixels of class $i$ predicted as class $j$
with $n_{cl}$ number of classes,
and let $t_i = \sum_j n_{ij}$ be the total number of pixels predicted as class $i$.
\figref{learning_curve} shows the loss, pixelwise accuracy,
and mean IU change with iterations:
the left side figures show the result with training dataset
and the right side ones show that with validation dataset.
This figure shows that the segmentation accuracies
are enhanced with both training and validation dataset, however,
after the iteration 6000 the mean IU with validation dataset decreases.
We used the trained network at training iteration 6000
for the actual prediction in the object segmentation task.
As for the segmentation accuracy at the iteration,
pixelwise accuracy is $0.906$ and mean IU is $0.283$ for validation dataset,
and this performance is as good as that by previous work \cite{long_shelhamer_fcn}
on a segmentation dataset NYUDv2 \cite{Silberman2012}: pixelwise accuracy $0.600$ and mean IU $0.292$.

% \figref{learning_curve}はは左から学習において与えたデータ数に対するロス，ピクセル単位での正解率，平均IUを示したグラフである．
% 上側の図の3つの図は学習データセットに対する学習曲線で，下の3つの図はテストデータセットに対するものを表す．
% この図から学習データセットとテストデータセットの両方で精度が向上していることがわかる．
% 実際にラベルを可視化した図でも\figref{learning_result}に示したように学習が進むにつれてセグメンテーション結果が向上していることがわかる．

% 4. Map generation for 3D Object Segmentation
% ## Map Generation
\section{Map Generation for 3D Object Segmentation} \seclab{octomap}

In this section, we introduce the function $M_t = g(M_{t-1}, I_t^{proba}, C_t)$
to generate 3D voxel grid map for object segmentation,
which consists of the registration of 2D probability map to 3D,
and map updates with the map representation with label occupancy grid.

% ### 2D確率画像の3次元点群への投影
\subsection{3D registeration of 2D probability map}

% 二次元の確率マップを三次元点群へ投影する方法
For extracting 2D object segmentation result three dimensionally,
the segmented region in the image is registered to the corresponding 3D point
using the correspondence relationship between image and point cloud.
This is a standard approach for the conversion of 2D image segmentation to 3D,
and in a occlusion-free environment enough number of 3D points of target object
are extracted with this operation for the shape and centroid estimation.
In environment with many occlusions, however,
it is hard to estimate the 3D shape and centroid with this
approach because of the lack of point cloud in some viewpoints.
It causes the displacement of the estimated centroid and different estimated size of the object.
Our method also includes this registration method, but we segment the target object
by generating a voxel grid map by sensing with different camera angles.
% 二次元画像における物体セグメンテーション結果を三次元情報として抽出する方法としては，
% 一般に二次元画像と三次元点群の対応関係が用いられ，
% 二次元情報を三次元点群の各ポイントに割り当てる(Registration)．
% これによって，二次元画像において目標物体であると判定された点群を抽出することが可能であり，
% 目標物体の三次元的な形状や重心点などを推定することができる．
% オクルージョンの多い環境においては，
% この結果を用いたタスク実現は物体点群の欠損の多さから困難で，
% 物体の一部しか点群が取れないことによる重心推定位置のずれや，
% 視点による物体の欠けによる推定サイズのずれが生じる．
% 本研究では，このような点群と物体確率の対応情報を複数の視点から獲得することで三次元マップを生成し，
% 三次元情報の欠けが多い環境においても適用可能な手法とする．

% 投影における問題点としてカメラが動き，Approximate Synchronizeでずれるが，
% これはロボットが止まっている場合にのみ更新することで解決．
Robot motion is required for the sensing of environment with different camera angles,
but there is displacement between RGB image and the point cloud sensed by the camera
when the camera is moving faster than a certain velocity.
This is because the RGB sensor and depth sensor activate in a different cycle,
and even with the synchronization of nearest two sensor input timestamps,
there is displacement in the index of pixels between RGB image and point cloud.
For this synchronization problem, we limited the map updates only when the camera is not moving,
and we guaranteed the correspondence between image pixels and depth points
with suspension of map updates until the velocity become 0 after camera movements.
% 複数の視点から環境をセンシングするためにはロボットの動作が必要となるが，
% 一定以上の速度でカメラリンクが移動している場合には一般に点群情報と画像情報にずれが生じる．
% これはRGB-Dカメラの画像センサと距離センサが異なる周期で動作しているためで，
% センシングのタイムスタンプによって最近傍のセンサ入力同士を同期させるようにした場合でも
% 画像と点群で物体の写るピクセル位置にずれが生じる．
% このため，本研究ではMapの更新のための投影はカメラリンクの速度が0である場合のみに限定し，
% 異なる視点へ移動した場合には速度が0になるまで更新を待つことで画像と点群の対応関係を保証した．

% ### 占有確率をグリッドによって更新するマップ更新法
\subsection{Map Generation with Label Occupancy Grid}

% 導入: What is occupancy grid map?
In this section, we introduce Label Occupancy Grid Map,
which is map updating method for object segmentation,
and the variation of Occupancy Grid Map \cite{30720},
where the environment is splitted into cells
with static grid size and the occupancy of each cell is represented probabilistically.
The map is usually used for path planning for mobile robot navigation \cite{30720} and
humanoid reaching behavior \cite{7487585},
but label occupancy grid map is map generation method for object segmentation
by replacing occupancy probability in original occupancy grid map with object probability.

% 確率の値の意味
In standard occupancy grid map,
the $i$-th cell $m_i$ holds the probability of occupancy $p(m_i)$.
$p(m_i)$ becomes close to 1 for the occupied cell,
close to 0 for the free cell, and around 0.5 for the unknown cell.
In our occupancy grid map, however, each grid probability means the occupancy by the target object.
For object segmentation, the object class is represented by label value with $0$ for background
and $1$ to $N$ for other objects.
So we call this occupancy grid map as Label Occupancy Grid Map,
in which each cell holds the probability of label occupancy $p_{lbl}(m_i)$:
$p(m_i) = 1$ means occupied by the label, $p(m_i) = 0$ means occupied by other label.
Our segmentation model includes background label and all pixel is labeled as each cell,
so $p(m_i) > 0.5$ means occupied by the label.

% 確率更新
We explain the method of updating $p(m_i)$ from sensor information $z_t$;
$t$ is the index of series of sensor input
which represents the period of the measurement time,
$z_t$ means the set of sensor information of $t$-th period,
and $z_{1:t}$ means the set of the sensor information
from the first period to the $t$-th period.
Notice that $z_t$ is set of the 3D point cloud and
the result of object probability prediction in 2D image.
By computing $p(m_i | z_{1:t})$ with $p(m_i | z_{1:t-1})$ and $z_t$,
we can integrate new sensor information and update the map periodically.

% 式
We can derive the following formula from Bayes' theorem and independence of the conditional probability.
\begin{eqnarray}
  p(m_i | z_{1:t})
  &=& \frac{p(z_t | m_i, z_{1:t-1}) \ p(m_i | z_{1:t-1})}{p(z_t | z_{1:t-1})} \nonumber\\
  &=& \frac{p(m_i | z_t) \ p(z_t) \ p(m_i | z_{1:t-1})}{p(m_i) \ p(z_t | z_{1:t-1})}
\end{eqnarray}
The ratio of occupied probability $p(m_i | z_{1:t})$ and free probability $p(\lnot m_i | z_{1:t})$ becomes simple as follows.
\begin{eqnarray}
  \frac{p(m_i | z_{1:t})}{p(\lnot m_i | z_{1:t})} = \frac{p(m_i | z_{1:t-1})}{p(\lnot m_i | z_{1:t-1})} \frac{p(m_i | z_t)}{p(\lnot m_i | z_t)} \frac{p(\lnot m_i)}{p(m_i)} \\
  {\rm where} \ \ \ \ p(\lnot m_i) = 1-p(m_i) \nonumber
\end{eqnarray}
We can deform the formula by introducing log-odds (logit) as follows.
\begin{eqnarray}
  l(m_i | z_{1:t}) = l(m_i | z_{1:t-1}) + l(m_i | z_t) - l(m_i)\\
  {\rm where} \ \ \ \ l(x) = {\rm logit}(p(x)) = \log \frac{p(x)}{1-p(x)} \nonumber
\end{eqnarray}
Presuming that we have no prior knowledge of the environment,
\begin{eqnarray}
  l(m_i) = 0 \ \ (\ \because \ \ p(m_i) = 0.5 \ )
\end{eqnarray}
Therefore,
\begin{eqnarray}
  l(m_i | z_{1:t}) = l(m_i | z_{1:t-1}) + l(m_i | z_t).
  \label{equ:update}
\end{eqnarray}
The probabilistic map is updated using this formula, and
the object probability $p(m_i | z_{1:t})$ is obtained
from the log-odds $l(m_i | z_{1:t})$ as follows:
\begin{eqnarray}
  p(m_i | z_{1:t}) = 1 - \frac{1}{1 + \exp l(m_i | z_{1:t})}
\end{eqnarray}

% 5. Experiment
% ## 実験
\section{Experiment}  \seclab{experiment}

\subsection{Experiment Setup}

% 実験環境
We validated our segmentation method with picking task experiments
using a life-sized humanoid robot
and octomap \cite{Octomap:AutonomousRobots2013:Hornung} as the efficient implementation of the
3D occupancy grid map.
To use original occupancy grid map for our label occupancy grid map,
we added the function of map updates using the object probability 2D map to the octomap.
The occupancy grid map is generated with 1[mm] resolution and the grids are
limited to the region inside a bin, with assumption that the shelf position and shape is known.
In the experiment, the robot localize the shelf by detecting the checkerboard,
and input point cloud to the shelf is segmented before being passed to the octomap
with known region clipping of the bin inside.
The given information about objects in the experiment is the target object and its located bin.
The set of candidate objects for object segmentation which can exist in the target bin
is unknown, so the object segmentation must be conducted with 40 classes exactly.

\subsection{Picking Task Execution based on Our Approach}

% 導入
We evaluated the method of object segmentation by generating the label occupancy map
with picking target objects from shelf bins.
\figref{hrp2_task_sequence} shows the sequential images which represents
the state of real robot (left side of each image) and visualization fo recognition (right side).
In the visualization, upper right image shows the RGB image input from camera which exists
in the head of the robot, lower right image shows the pixelwise probability of target object,
and left region shows the point cloud and generated 3D map.

% タスクシーケンス
The task is conducted as follows: firstly robot looks around the target bin to generate label
occupancy grid map, secondly picks the object based on the generated map,
and finally places the object at the ordered place.
The camera poses with the looking around motion by robot are decided by the target bin and
given shelf geometry and its pose with a checkerboard,
setting fixed offset from shelf to the standing point and
fixed 3 different camera angles in 4 different heights of standing.
If the voxels for target object are generated with this recognition stage,
the picking motion of humanoid robot to the object is generated based on depth value of the
centroid of computed from the generated 3D voxels afterwards.
There can be some optimization in looking around motion,
but the result with this motion shows the ability of generating object voxels
even with few good views for the target object by using the few good views.

% 実験結果
We conducted the picking task for 3 objects (t-shirts, cup and dumbbell) from 3 bins.
\figref{hrp2_task_sequence} shows the picking sequence for t-shirts,
and \figref{transition_graph_iu} shows the segmentation accuracy and camera velocity
while the looking around motion.
The segmentation accuracy is measured with IU between the 3D box annotated by human
and generated voxels. Here, IU is defined as $V_{tp} / V_{tp} + V_{fp} + V_{fn}$:
$V_{tp}$ is overlap volume between the box and voxels (true positive),
$V_{fp}$ is difference between $V_{tp}$ and voxels volume (false positive),
$V_{fn}$ is one between $V_{tp}$ and box volume (false negative).
\figref{transition_graph_iu} shows the rise of accuracy (IU) with different camera views.
From the transition of camera velocity it is supposed that IU after first view is around 0.05
at time $1 [sec]$, and after 4 views (at time $12 [sec]$)
the IU is around 0.11 and it increases twofold.
These result shows our proposed method is effective to segment object densely using multiple views.
With the task sequence, the robot successfully picked the 3 objects from shel bin and placed
them into the ordered place.
% \figref{hrp2_pick_dumbbell} and \figref{hrp2_pick_cup}
% shows the picking sequence for the dummbel and cup.
% The lower images in \figref{hrp2_pick_dumbbell} shows the frames from inside the bin,
% and it shows the robot reach the arm correctly to the object center, in the condition
% where it is likely the robot reach the front side of the object center without our method.
% \figref{hrp2_pick_cup} shows the harder situation to pick the object, because the cup
% is located the deep position in the bin.
% Even in this harder situation for picking, the robot reached the hand to
% the center of the object and successfully picked it.

\begin{figure*}[t]
  \centering
    \subfloat[]{
      \includegraphics[height=3.3cm]{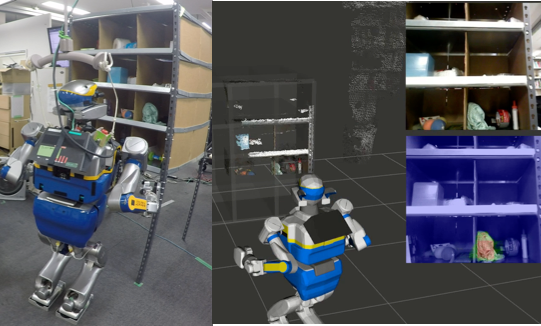}
      \figlab{exp00001}}
    \subfloat[]{
      \includegraphics[height=3.3cm]{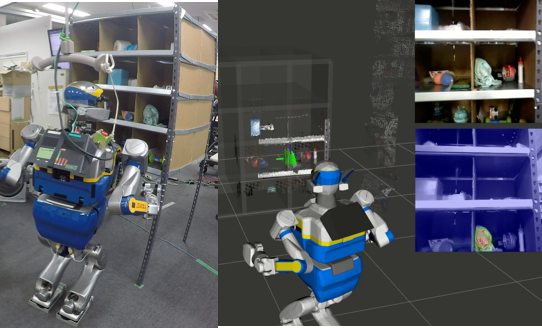}
      \figlab{exp00002}}
    \subfloat[]{
      \includegraphics[height=3.3cm]{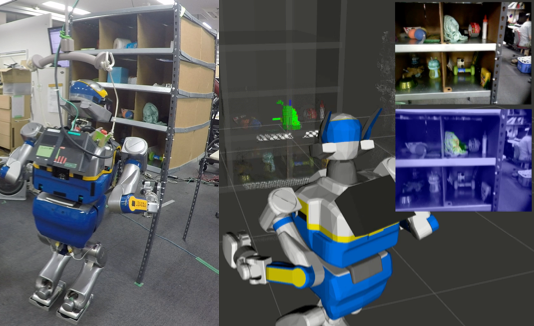}
      \figlab{exp00003}}

    \subfloat[]{
      \includegraphics[height=2.95cm]{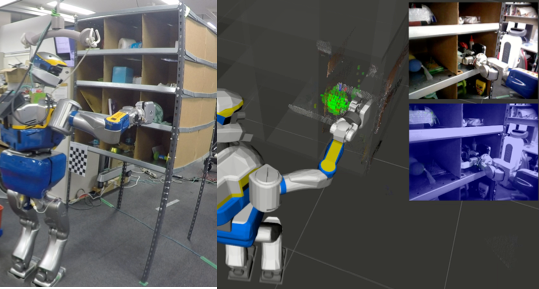}
      \figlab{exp00004}}
    \subfloat[]{
      \includegraphics[height=2.95cm]{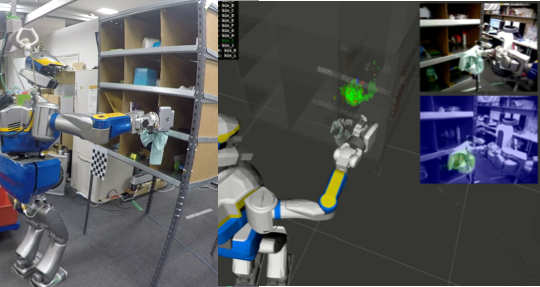}
      \figlab{exp00005}}
    \subfloat[]{
      \includegraphics[height=2.95cm]{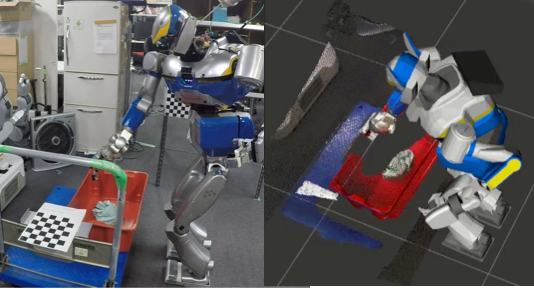}
      \figlab{exp00006}}
  \caption{Task sequence of robot picking t-shirts from a bin and place to the ordered bin. \newline
    \footnotesize{
      The left side of each image represent the state of real robot, and the right side shows
      the result of visualization.
      In the experiment, the robot segment target object with looking around motion,
      and reach the hand to the centroid of generated 3D object map.
    }
  }
  \figlab{hrp2_task_sequence}
\end{figure*}

\begin{figure}[htbp]
  \centering
  \subfloat[]{
    \includegraphics[width=0.44\textwidth]{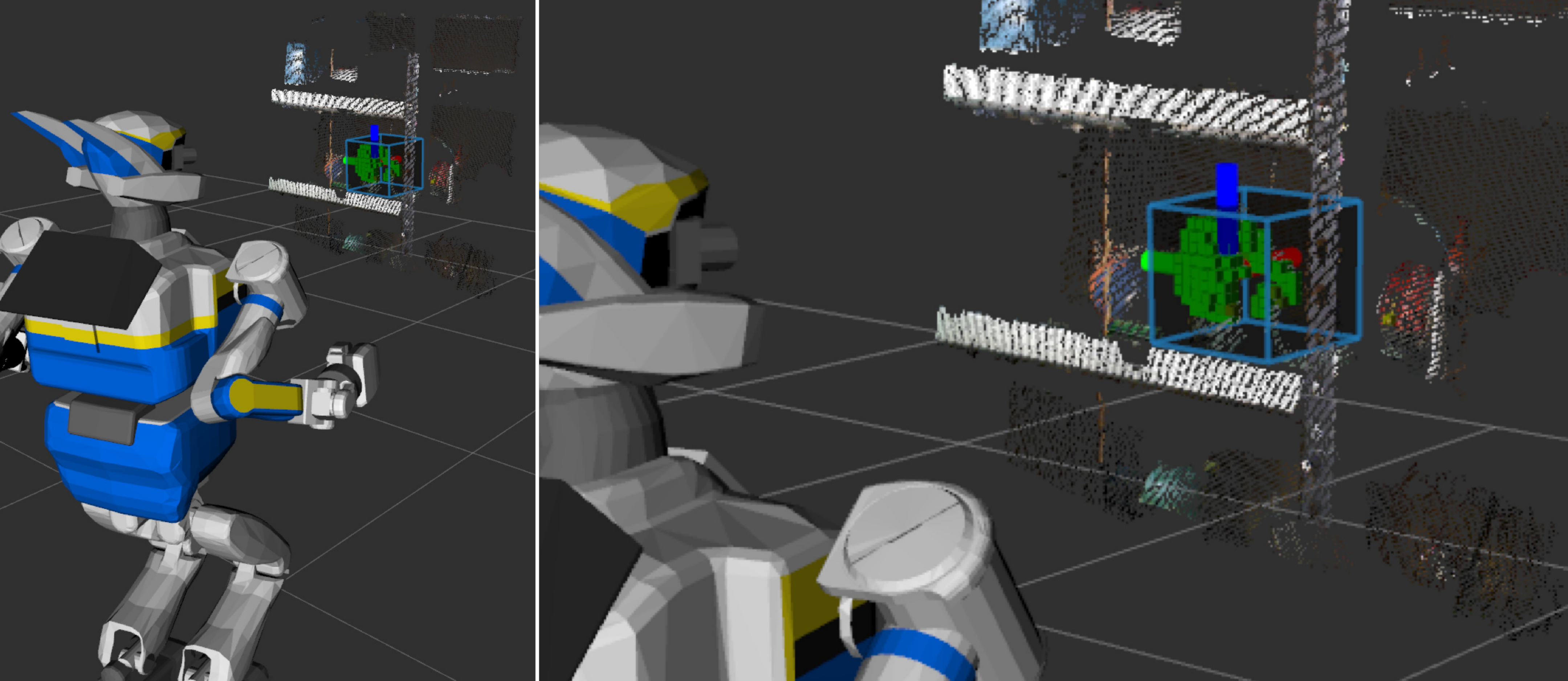}
    \figlab{human_annotation}
  }

  \vspace*{-3mm}
  \hspace*{-3mm}
  \subfloat[]{
    \includegraphics[width=0.48\textwidth,trim={5mm 1mm 5mm 8mm},clip]{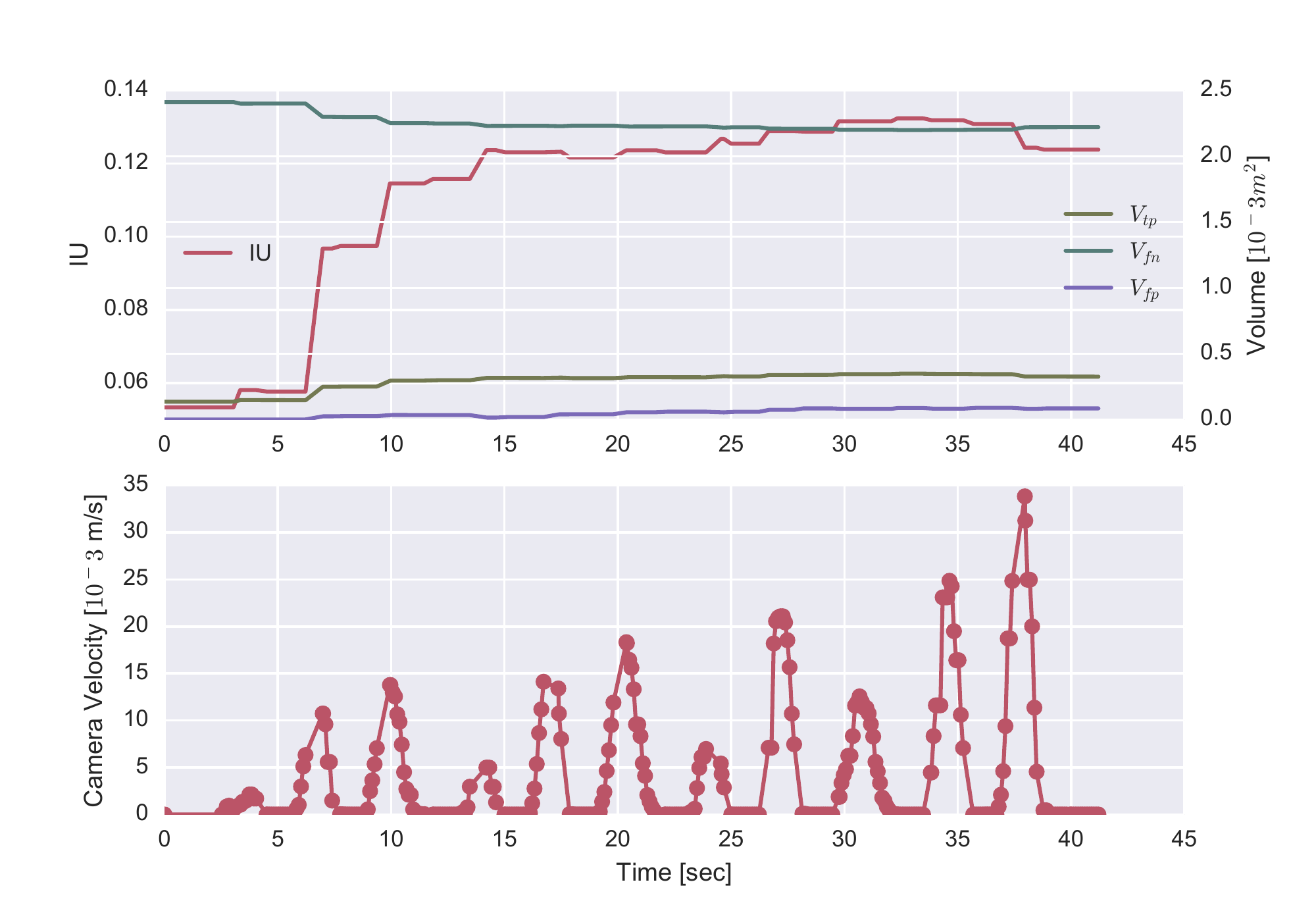}
    \figlab{graph_iu_2}
  }

  % \subfloat[]{
  %   \includegraphics[width=0.48\textwidth,trim={5mm 1mm 5mm 8mm},clip]{live_20160927_eval_label_12}
  %   \figlab{graph_iu_12}
  % }
  %
  % \subfloat[]{
  %   \includegraphics[width=0.48\textwidth,trim={5mm 1mm 5mm 8mm},clip]{live_20160927_eval_label_16}
  %   \figlab{graph_iu_16}
  % }
  \caption{Transition graph of segmentation accuracy and camera velocity
    while looking around to pick t-shirts.
    \newline
    \footnotesize{
      Horizontal axis indicates the time as an elapsed time in second,
      and vertical axis does IU, volumes, and camera velocity.
      The accuracy is computed as the IU between voxels and 3D box annotated by human
      as shown in \figref{human_annotation}.
    }
  }
  \figlab{transition_graph_iu}
\end{figure}

% \begin{figure}[htbp]
%   \centering
%   \subfloat[]{
%     \includegraphics[height=3.2cm]{experiment_dumbbell_001}
%     \figlab{experiment_dumbbell_001}
%   }
%   \subfloat[]{
%     \includegraphics[height=3.2cm]{experiment_dumbbell_002}
%     \figlab{experiment_dumbbell_002}
%   }
%   \subfloat[]{
%     \includegraphics[height=3.2cm]{experiment_dumbbell_003}
%     \figlab{experiment_dumbbell_003}
%   }
%   % \includegraphics[width=0.45\textwidth]{experiment_dumbbell_frames}
%   % \includegraphics[width=0.5\textwidth]{experiment_dumbbell_frames_in_bin}
%   \caption{Task sequence of robot picking dumbbell from a bin.}
%   \figlab{hrp2_pick_dumbbell}
% \end{figure}

% \begin{figure*}[htbp]
%   \centering
%   \subfloat[]{
%     \includegraphics[height=3.2cm]{experiment_cup_001}
%     \figlab{experiment_cup_001}
%   }
%   \subfloat[]{
%     \includegraphics[height=3.2cm]{experiment_cup_002}
%     \figlab{experiment_cup_002}
%   }
%   \subfloat[]{
%     \includegraphics[height=3.2cm]{experiment_cup_003}
%     \figlab{experiment_cup_003}
%   }
%   % \includegraphics[width=0.45\textwidth]{experiment_cup_frames.png}
%   \caption{Task sequence of robot picking cup from a bin. \newline
%     \footnotesize{
%       The cup is placed the deep position in a bin,
%       and it is difficult to sense the point cloud of that object without our mapping method.
%     }
%   }
%   \figlab{hrp2_pick_cup}
% \end{figure*}

% 6. Conclusions
\section{CONCLUSIONS}

We presented an approach to segment object three dimensionally in a narrow space,
acquiring dense 3D information with deep learning and voxel grid map.
The contributions of this paper are the following:
\begin{enumerate}
  \item we proposed a novel approach to segment object three dimentionally with integration of
    deep learning and occupancy voxel grid map.
  \item we evaluated the effectiveness our approach with shelf picking task by life-sized humanoid.
\end{enumerate}
By introducing map generation method which uses object probability,
we extended the applicability of object segmentation
in environment with lack of point cloud of objects because of occlusions.
We verified the effectiveness of our method with experiments on
picking task execution for target objects in narrow bins.
With our method, the object 3D model is generated densely in an environment with occlusions,
and picking task is completed successfully using the centroid estimation.
For future work, it is considered that our method is usable for object exploration
by updating the label occupancy grid map until the object is detected.

% \addtolength{\textheight}{-12cm} % This command serves to balance the column lengths
                                   % on the last page of the document manually. It shortens
                                   % the textheight of the last page by a suitable amount.
                                   % This command does not take effect until the next page
                                   % so it should come on the page before the last. Make
                                   % sure that you do not shorten the textheight too much.

% \input src/appendix.tex
% \input src/acknowledge.tex

\bibliographystyle{junsrt}
\bibliography{main}

\end{document}